\documentclass[10pt,journal]{IEEEtran}

\ifCLASSINFOpdf
\else
\fi
\usepackage{times}
\usepackage{epsfig}
\usepackage{graphicx}
\usepackage{amsmath}
\usepackage{amssymb}

\usepackage{footnote}
\usepackage{multirow}
\usepackage{booktabs}
\usepackage{threeparttable}
\usepackage{enumerate}
\usepackage{indentfirst}
\usepackage{array}
\usepackage{rotating}
\usepackage{ragged2e}
\usepackage{wasysym}
\usepackage{caption}
\usepackage{epstopdf}
\usepackage{color}
\usepackage{algorithm}
\usepackage{times}
\usepackage{epsfig}
\usepackage{graphicx}
\usepackage{amsmath}
\usepackage{amssymb}
\usepackage{epstopdf}
\usepackage{color}
\usepackage{array}
\usepackage{tabu} 
\usepackage{comment}
\usepackage{algorithm}
\usepackage{algorithmicx}
\usepackage{cases}
\usepackage{bm}
\usepackage{bbm}
\usepackage{algpseudocode}
\usepackage{booktabs}
\usepackage{multirow}
\usepackage{textcomp}

\def\eg{{\em e.g.}}
\def\etal{{\em et al.}}

\newcommand{\secref}[1]{Section \ref{#1}}

\newcommand{\br}[1]{\textbf{#1}}

\begin{document}
	%
	\title{A Sparse Representation Based Joint Demosaicing Method for Single-Chip Polarized Color Sensor}
	%
	%
	%
	
	\author{Sijia~Wen,
		Yinqiang~Zheng,~\IEEEmembership{Member,~IEEE}
		and~Feng~Lu,~\IEEEmembership{Member,~IEEE}
		\IEEEcompsocitemizethanks{
			\IEEEcompsocthanksitem Manuscript received June 08, 2020; revised February 16, 2021; accepted March 20, 2021. This work was supported by National Natural Science Foundation of China (NSFC) under Grant 61732016. (\emph{Corresponding author: F. Lu and Y. Zheng})\protect
			\IEEEcompsocthanksitem S. Wen and F. Lu are with the State Key Laboratory of Virtual Reality Technology and Systems, School of Computer Science and Engineering, Beihang University, Beijing 100191, China. Email: $\{$sijiawen, lufeng$\}$@buaa.edu.cn\protect
			\IEEEcompsocthanksitem Y. Zheng is with the Next Generation Artificial Intelligence Research Center, The University of Tokyo, Tokyo 113-8656, Japan. E-mail: yqzheng@ai.u-tokyo.ac.jp. \protect
			\IEEEcompsocthanksitem S. Wen and F. Lu are also with Peng Cheng Laboratory, Shenzhen, China. \protect	}
	}

	\maketitle
	
	\begin{abstract}
		The emergence of the single-chip polarized color sensor now allows for simultaneously capturing chromatic and polarimetric information of the scene on a monochromatic image plane. However, unlike the usual camera with an embedded demosaicing method, the latest polarized color camera is not delivered with an in-built demosaicing tool. For demosaicing, the users have to down-sample the captured images or to use traditional interpolation techniques. Neither of them can perform well since the polarization and color are interdependent. Therefore, joint chromatic and polarimetric demosaicing is the key to obtaining high-quality polarized color images. In this paper, we propose a joint chromatic and polarimetric demosaicing model to address this challenging problem. Instead of mechanically demosaicing for the multi-channel polarized color image, we further present a sparse representation-based optimization strategy that utilizes chromatic information and polarimetric information to jointly optimize the model. To avoid the interaction between color and polarization during demosaicing, we separately construct the corresponding dictionaries. We also build an optical data acquisition system to collect a dataset, which contains various sources of polarization, such as illumination, reflectance and birefringence. Results of both qualitative and quantitative experiments have shown that our method is capable of faithfully recovering full RGB information of four polarization angles for each pixel from a single mosaic input image. Moreover, the proposed method can perform well not only on the synthetic data but the real captured data.
	\end{abstract}
	

	%
	\IEEEpeerreviewmaketitle

	\section{Introduction}\label{sec:intro}
	
	\IEEEPARstart{C}{onventional} color imaging can sample spectral information. Polarization imaging considers the electric field as a vector which is contained in a plane perpendicular to the direction of propagation. It is a way to analyze the particular direction of the oscillation of the electric field described by the light. Color is easily perceptible to human eyes and is at the core of numerous computer vision problems. On the contrary, polarization is invisible to our eyes, yet it usually conveys critical information on intrinsic material properties, such as reflectance and external surface geometry. For the color cameras, various approaches~\cite{mairal2009non,getreuer2011zhang,heinze2012joint,wang2014multilayer} have been proposed for chromatic demosaicing. Similarly, several polarimetric demosaicing methods~\cite{zhang2018sparse,zhang2018learning,gao2013gradient,zhang2016image,ratliff2009interpolation,gao2011bilinear} have been proposed to reconstruct full resolution polarization images.
	\begin{figure}[htb]
		\begin{center}
			\includegraphics[width=1\linewidth]{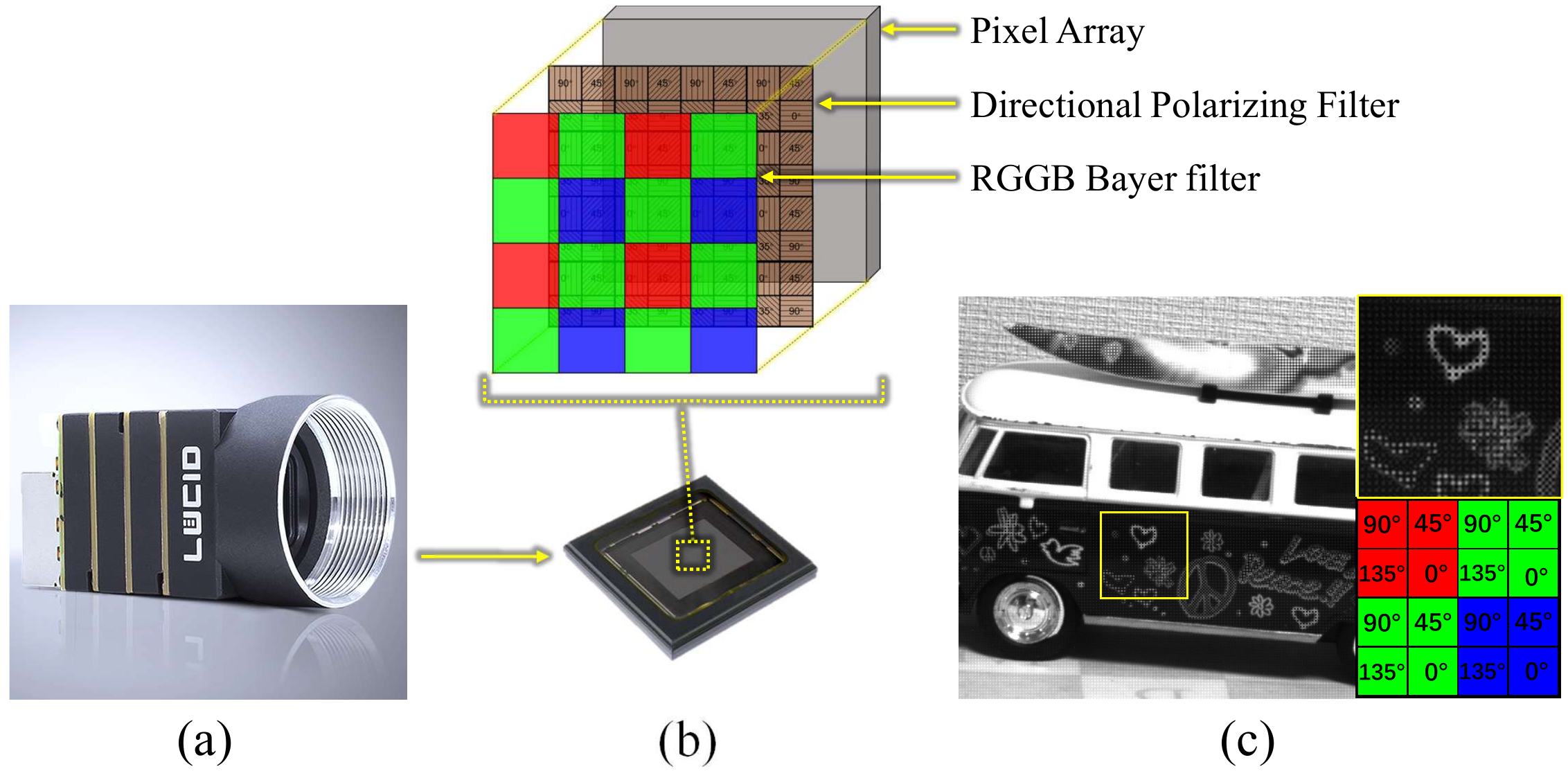}
		\end{center}
		\caption{The polarized color camera: (a) The single-chip polarized color camera by Lucid Vision, which is equipped with a Sony IMX250MYR CMOS sensor; (b) The polarized color sensor with an RGGB Bayer filter~\cite{bayer1976color} in addition to the directional polarizing filter to form a $4 \times 4$ array; (c) The RGB-Polarization pattern of IMX250MYR which can simultaneously capture chromatic information and polarimetric information in the mosaic form.}
		\label{pattern}
	\end{figure}
	
	Previously, we could only get either chromatic information or polarimetric information of the scene by each snapshot. Until recently, the newly released polarized color cameras (\eg FLIR BFS-U3-15S5PC and PHX050S camera) are able to simultaneously capture RGB and polarization pixels of the scene in the mosaic form. These cameras are equipped with an RGGB Bayer filter and the directional polarizing filter to form a 16 pixels calculation unit as the RGB-Polarization pattern, as shown in Fig.~\ref{pattern}. Leveraging on the polarized color camera, the full resolution polarized color images can be a benefit to a great variety of computer vision tasks~\cite{wen2021polarization}. For example, the LCD panels or mobile phone displays can benefit from reflection removal while maintaining the shape and color integrity of the object in the image; Using such a camera can also ease the procedure of sugar concentrations measuring in industry, as shown in \secref{realimage}. 
	
	Since demosaicing is the first step of polarized color image processing pipeline and is the key to achieving full resolution polarized color images. Unlike chromatic demosaicing or polarimetric demosaicing, the goal of the polarized color image demosaicing is to recover the full 12 channels ($(r, g, b)*(0^\circ, 45^\circ, 90^\circ, 135^\circ)$ from the one channel mosaic image. Therefore, the traditional interpolation methods cannot perform well on polarized color mosaic images. Moreover, due to the complicated interdependent correlation between chromaticity and polarization, obtaining satisfactory results is still a challenging problem for previous demosaicing methods. 
	
	In this paper, we propose a sparse representation-based joint chromatic and polarimetric demosaicing method to restore high-quality 12-channel images from the 1-channel mosaic observations, which are captured by the single-chip polarized color sensor. By observing the polarized color images, we find that the object color might change due to the different polarization angles. This reveals the interaction of color and polarization for light reflected from an opaque surface. In turn, this phenomenon necessitates the joint chromatic and polarimetric demosaicing algorithm to account for this interaction. To consider the correlation between chromaticity and polarization in the polarized color image, we first build a joint demosaicing model to reformulate the task of polarized color image demosaicing by jointing the chromatic information and polarimetric information. Then, to avoid the mutual interference of color and polarization information during the demosaicing, we separately construct the chromatic and polarimetric dictionaries by corresponding signal data. At last, we design a customized ADMM~\cite{boyd2011distributed} optimization scheme with sparse coding to implicitly reconstruct the chromatic information and polarimetric information. The key point of the proposed method is jointly recovering the missing information but separately constructing the dictionaries. We further impose the polarimetric constraint on the joint demosaicing model to ensure that the chromatic information and the polarimetric information will not affect each other.
	
	To construct the dictionaries, we build a practical optical data acquisition system to collect a dataset. In the ideal reflection scene, the color of the object should be similar under different polarization angles. However, multiple sources of polarization, such as illumination and birefringence, will make the color of the scene vary with different polarization angles. To accurately capture the RGB information of the scene under four polarization angles, we use a polarizer, an RGB filter, a mono camera, motorized rotators to capture each R, G, B information of four polarization angles separately. In addition, we also collect a real captured dataset to evaluate the performance in the wild. Experimental results have shown that our proposed methods are capable of faithfully recovering full-resolution RGB images for all four angles of polarization from a single mosaic image.
	
	Our main contributions are summarized as follows: 1) We propose a joint demosaicing model for the polarized color image demosaicing; 2) We design a sparse representation-based optimization strategy for the cross interpolation of RGB and polarimetric information; 3) We build a data acquisition system to collect an RGB-Polarization dataset, which contains many sources of polarization, like illumination and birefringence;  and 4) We conduct extensive experiments and demonstrate that our proposed method achieves state-of-the-art results in terms of quantitative measures and visual quality.
	
	The rest of this paper is organized as follows: \secref{related work} reviews related work and \secref{Methodology} presents the methodology of the proposed approach. \secref{optimization} details the scheme of the optimization and \secref{experiments} demonstrate the experiments. Finally, \secref{conclution} concludes the paper.
	
	\section{Related work}\label{related work}
	Since image demosaicing is an important and indispensable problem, plenty of researches has been reported and most of them focus on either image quality improvement or applying it to specific applications~\cite{li2008image,kaur2015survey}. They can be further divided into the chromatic demosaicing, the polarimetric demosaicing, and joint chromatic and polarimetric demosaicing. In this section, we overview all these categories as related works.
	
	\subsection{Chromatic demosaicing}The existing commodity color camera records mosaiced RGB images on a single monochromatic senor by using the color filter arrays (CFAs~\cite{menon2011color}), like the well-known Bayer pattern~\cite{bayer1976color}. Since the CFA technology is lightweight, cheap, robust, and small enough to be embedded in imaging systems, the technology has quickly become the standard for one-shot color imaging. It is composed of a single silicon sensor fitted with a CFA, so that each sensor site senses only one spectral band according to the CFA. The chromatic demosaicing aims to reconstruct 3 channels ($r, g, b$) from the one channel RGB mosaic image, which recorded by the CFA. 
	
	Some of the traditional chromatic demosaicing algorithms rely on the statistical formulation in the spatial domain. Those works can be summarized into a constant-hue assumption made by the majority of existing demosaicing algorithms~\cite{li2005demosaicing,zhang2005color,jaiswal2014exploitation}. Since the corresponding low-pass filters can suppress the aliasing by eliminating overlapped high-frequency components, methods~\cite{alleysson2005linear,dubois2005frequency,lian2007adaptive} recover the missing information by frequency domain analysis. By taking advantage of spatial self-similarities, methods~\cite{buades2009self,zhang2011color} use non-local self-similarity priors as regularization terms to enhance the performance of an interpolation model. In order to fully use the information of images, optimization schemes~\cite{condat2012joint,klatzer2016learning,kiku2016beyond}, and compressive sensing~\cite{mairal2009non,moghadam2013compressive,dave2017compressive} are proposed to obtain more accurate results.
	
	More recently, demosaicing strategies based on neural networks lead to better quality and efficiency. Kappa \etal~\cite{kapah2000demosaicking} and Go \etal~\cite{go2000interpolation} are among the first to use neural networks for color demosaicing. Long \etal~\cite{long2006adaptive} later proposed an adaptive scheme to improve performance. Heinze \etal~\cite{heinze2012joint} proposed a multi-frame demosaicing algorithm using a neural network to infer the pixel color based on its surroundings. Wang~\cite{wang2014multilayer} used ${4 \times 4}$ patches to train a multilayer neural network while minimizing a suitable objective function. Gharbi \etal~\cite{gharbi2016deep,henz2018deep} constructed a dataset with hard cases, which were used to train a CNN for joint demosaicing and denoising. Heide \etal~\cite{heide2016proximal} later organized the principles of algorithm design in FlexISP into ProxImaL, a domain specific language for optimization based image reconstruction. Chen \etal~\cite{chen2018learning} provides high-quality RGB images from single raw images taken under low-light conditions.
	
	\subsection{Polarimetric demosaicing}Snapshot polarization imaging has gained interest in the last few decades. Recent research and technology achievements defined the polarization Filter Array (PFA~\cite{tokuda2009polarisation}). Polarization Filter Array imaging provides a way for snapshot acquisition that could be useful for many computer vision tasks. The PFA is composed of pixel-size linear polarizers oriented at four different angles ($0^\circ, 45^\circ, 90^\circ, 135^\circ$) superimposed on a \label{key}rcamerasor chip. The goal of polarimetric demosaicing is to perform on each sparse channel to obtain an estimated image with four fully-defined channels, among which three are estimated at each pixel. Snapshot polarization imaging has gained popularity due to the recent advancements in producing micro DoFP polarimeters, with successful applications in analyzing the light electric field oscillation direction~\cite{mihoubi2018survey}, material classification~\cite{hu2016polarization}, 3-D surface reconstruction~\cite{drouet20143d}, dehazing~\cite{fang2014image}, and biomedical imaging~\cite{alali2015polarized}.
	
	A few methods have been proposed to address the demosaicing issue in the polarimetric domain. Bilinear interpolation was first investigated by Ratliff \etal~\cite{ratliff2009interpolation} who also proposed an extension of the bilinear interpolation. Tyo \etal~\cite{tyo2009total} developed a new method to reconstruct the first three Stokes vector directly from the mosaicked image. Zhang \etal\cite{zhang2016image} took advantage of the correlations in PFA (polarimetric filter array) to enhance the spatial resolution. Moreover, a new interpolation method for DoFP imaging sensors with intensity correlation was presented in~\cite{ahmed2017residual}. A demosaicing model based on sparse representation is proposed in~\cite{zhang2018sparse}. A customized polarimetric demosaicing convolutional neural network (PDCNN) was recently presented in~\cite{zhang2018learning}.
	
	\subsection{Joint Chromatic and Polarimetric demosaicing}Since the polarized color image demosaicing aims to estimate 12 channels ($(r, g, b)*(0^\circ, 45^\circ, 90^\circ, 135^\circ)$ from the one channel polarized color mosaic image, neither chromatic demosaicing nor polarimetric demosaicing can apply to the newly released RGB-Polarization pattern. Due to the numerous missing information, the baseline interpolation algorithm cannot obtain promising results. 
	
	Recently, based on the polarized color camera, the newly published CPDNet~\cite{wen2019convolutional} uses the captured data as ground truth and the corresponding synthesized mosaic images as input data to train the network. However, collecting data in their way will bring unexpected chromatic aberration, caused by a prism-based RGB camera. In addition, the dataset collected by Wen~\cite{wen2019convolutional} didn't contain the polarization from birefringence, which will decrease the performance on real scenes. To this end, different from CPDNet, we build a data acquisition system with a mono camera to collect accurate data. Instead of fitting the synthesized dataset, we transform the joint demosaicing problem into an energy function that can be solved by the sparse representation based ADMM optimization scheme. Qiu~\etal~\cite{qiu2019polarization} propose a polarization demosaicking algorithm by inverting the polarization image formation model for both monochrome and color DoFP cameras. However, their model is based on the Stokes vector without reconstructing the polarized color image. In this case, they can only reconstruct the good polarization parameter results instead of full-resolution polarized color images. In addition, their demosaicing algorithm ignores the interdependent correlation between polarization and chromaticity.
	
	\section{Methodology}\label{Methodology}
	
	\subsection{Problem Statement}
	Traditional interpolation estimates the missing information based on the analysis of the input observation. However, due to the sparse of each channel of the polarized color mosaic image, the traditional interpolation algorithm cannot recover the information truthfully. The former dictionary-based methods are well known for color or polarimetric demosaicing. However, to use them in a mechanical way will not suffice here, especially for some surface materials with complex light transport properties. Deep learning-based methods infer the color and polarization pixels by fitting the dataset. Yet the RGB information and polarimetric information will be mixed, and impact each other during the training. In addition, the training phase of deep learning methods consumes a lot of time. 
	
	Inspired by the previous demosaicing methods based on sparse representation~\cite{zhang2018sparse,mairal2009non,huang2014dictionary}, we propose a sparse representation-based joint demosaicing method for single-chip polarized color sensor. Different from the demosaicing method for the chromatic image or polarimetric image, the joint chromatic and polarimetric demosaicing need to recover the missing pixels from one out of twelve necessary intensity measurements. For each channel $\theta$, the observed image $\br{I}$ is essentially down-sampled from its full-resolution image $\br{Y}$. 
	\begin{equation} \label{1}
	\br{I} = \sum_{\theta}Mask_\theta \br{Y}.
	\end{equation}
	where $\theta = {(r, g, b)*(0^\circ, 45^\circ, 90^\circ, 135^\circ)}$. $Mask_\theta$ represent down-sample matrix based on the RGB-Polarization pattern. 
	
	In order to take advantage of the interaction among different channels, the polarized color image at 12 channels should be joined to build the demosaicing model. Meanwhile, to prevent the chromaticity and polarization from affecting each other, we need to recover the missing information of chromaticity and polarization separately. In this regard, we can take advantage of the sparse representation-based method to address this issue. According to sparse representation theory, the joint demosaicing for the polarized color camera can be transformed to minimize the following problem:
	\begin{equation} \label{total1}
	\begin{aligned}
	\mathop{min}\limits_{\br{D}_\theta,\br{X}_\theta} &\{\sum_{\theta}\|\br{I} - Mask_\theta \br{Y}_\theta\|_F^2 +\\
	&\sum_{\theta}\lambda_\theta \|\br{Y}_\theta - \br{D}_\theta \br{X}_\theta\|_F^2 + \rho \|\br{X}_\theta\|_1 \}.
	\end{aligned}
	\end{equation}
	where $\br{D}_\theta, \br{X}_\theta$ are the dictionary and the corresponding sparse coding. $\lambda, \rho$ are the parameters to balance the effect of different channels. 
	
	Sparse representation is essentially composed of dictionary learning and sparse coding. It seems like the problem can directly be solved by learning the dictionary of each channel ($\br{D}_\theta$) and calculating the corresponding sparse coding ($\br{X}_\theta$). However, for some surface materials with complex light transport properties, the full-resolution image might show different colors at different polarizing status. The most obvious example is birefringence, for which an optically anisotropic material has a refractive index that depends on the polarization and the prorogation direction of light. In other words, there is a non-negligible correlation between each channel, so $\lambda_\theta$ cannot be set to a static value. Taking the entire input data as the dictionary is computationally inaccurate and consumes too much storage space. In addition, constructing one dictionary from both RGB information and polarimetric information ignores their mutual interference. The experimental results in Tab.~\ref{comparison3} verify our theory.
	
	Therefore, the correlation between different RGB and polarization channels is key to addressing this interpolation issue. In this paper, we build a joint demosaicing model to address this challenging problem. 
	
	\subsection{Joint Demosaicing Model}
	As we mentioned above, to exploit the correlation between chromaticity and polarization without affecting each other, we need to jointly reconstruct the RGB information and polarimetric information with different constructed dictionaries. In this case, we reformulate the joint demosaicing problem to minimize the following energy function:
	\begin{equation} \label{model}
	\begin{split}
	\mathop{min}\limits_{\br{RC},\br{RP}} &\|\br{I} - Mask_{rgb}\br{RC} - Mask_{pol}\br{RP}\|_F^2 +\\
	&\Phi(\br{RP})+ \Psi(\br{RC}).
	\end{split}
	\end{equation}
	where $\br{I}$ is the input polarized color mosaic image. $Mask_{rgb}$ and $Mask_{pol}$ represent the down-sample matrices of the reconstructed RGB information $\br{RC}$ and reconstructed polarimetric information $\br{RP}$, respectively. The first term aims to maintain the accuracy between the input and reconstructed images, which jointly build the interpolation model. The latter two items designate the implicit priors imposed on $\br{RP}$, $\br{RC}$ to regularize inference. $\Phi(\br{RP})$ and $\Psi(\br{RC})$ separately calculate the expectant reconstructed results of the polarization and chromaticity via sparse coding, which avoid the interaction of color and polarization during the demosaicing.
	
	Then we adopt the ADMM optimization scheme to recover the missing pixels in RGB and polarization channels. As the well-known idea of ADMM, we need to introduce the auxiliary variables $\br{P}, \br{C}$ to Eq.~\ref{model} to constrain the representation of $\br{RP}$ and $\br{RC}$. Then the Eq.~\ref{model} can be formulated as the following optimization problem with both chromatic and polarimetric constraints:
	\begin{equation} \label{model1}
	\begin{split}
	\mathop{min}\limits_{\br{RC},\br{RP},\br{C},\br{P}} &\|\br{I} - Mask_{pol}\br{RP} - Mask_{rgb}\br{RC}\|_F^2+\\
	&\br{S}_{pol}^T(\br{RP}-\br{P}) + \frac{\rho_{pol}}{2}\|\br{RP}-\br{P}\|_F^2 +\\ &\br{S}_{rgb}^T(\br{RC}-\br{C}) + \frac{\rho_{rgb}}{2}\|\br{RC}-\br{C}\|_F^2.
	\end{split}
	\end{equation}
	where $\br{S}_{pol}, \br{S}_{rgb}$ and $\rho_{pol}, \rho_{rgb}$ are multipliers and Lagrangian penalty parameters for $\br{RC},\br{RP}$. $\br{P}$ and $\br{C}$ are the desire results of $\br{RP}$ and $\br{RC}$. By introducing the scaled Lagrange multipliers:
	\begin{equation} \label{multipliers}
	\begin{split}
	y_{pol}^{k} &= (1/\rho_{pol})\br{S}_{pol}^{(k)},\\
	y_{rgb}^{k} &= (1/\rho_{rgb})\br{S}_{rgb}^{(k)}.
	\end{split}
	\end{equation}
	Eq.~\ref{model1} can be expressed as:
	\begin{equation} \label{sub1}
	\begin{split}
	\mathop{min}\limits_{\br{RC},\br{RP},\br{C},\br{P}} &\|\br{I} - Mask_{pol}\br{RP} - Mask_{rgb}\br{RC}\|_F^2+\\
	\frac{\rho_{pol}}{2}&\|\br{RP}-(\br{P}^{k} - y_{pol}^{k})\|_F^2 +\\ \frac{\rho_{rgb}}{2}&\|\br{RC}-(\br{C}^{k} - y_{rgb}^{k})\|_F^2.
	\end{split}
	\end{equation}
	
	To minimize Eq.~\ref{sub1}, we need to calculate the sub-problems about $\br{P}$, $\br{C}$, which can be written as:
	\begin{equation} \label{sub2}
	\begin{cases}
	\br{P}^{k+1} = \arg\mathop{min}\limits_{\br{P}}f(\br{P}) + \frac{\rho_{pol}}{2}\|\br{P} - (\br{RP}^{k} + y_{pol}^{k})\|_F^2,\\
	\br{C}^{k+1} = \arg\mathop{min}\limits_{\br{C}}g(\br{C}) + \frac{\rho_{rgb}}{2}\|\br{C} - (\br{RC}^{k} + y_{rgb}^{k})\|_F^2.
	\end{cases}
	\end{equation}
	where $f(\br{P})$ and $g(\br{C})$ are the implicit functions imposed on $\br{P}$, $\br{C}$. By taking advantage of sparse representation, we separately calculate $f(\br{P})$ and $g(\br{C})$. The detail of these calculations is presented in \secref{ini}.
	
	\section{Optimization}\label{optimization}
	It found that the optimization can be iterated only if we obtained the desire value of variables $\br{P}, \br{C}$. Therefore, we define the $f(\br{P}), g(\br{C})$ with sparse coding and dictionary learning to obtain the approximate results for $\br{RP}$, $\br{RC}$, respectively. 
	
	\subsection{Constructing the Dictionaries}
	As we mentioned above, $f(\br{P}), g(\br{C})$ aim to obtain the desire reconstruction results of polarimetric and RGB information without the mutual interference of color and polarization. The performance of $f(\br{P}), g(\br{C})$ plays a key role during the iteration of optimization. Inspired by~\cite{aharon2006k,dumitrescu2017regularized,elad2006image}, sparse coding and dictionary learning can help to address this issue. In sparse and low-rank representations, constructing a proper dictionary is important. We use the same strategy to construct the polarimetric dictionary $\br{D}_{pol}$ and the chromatic dictionary $\br{D}_{rgb}$. The difference between them is how to generate the corresponding signal data extracted from the RGB-Polarization dataset. 
	\begin{figure}[b]
		\begin{center}
			\includegraphics[width=1\linewidth]{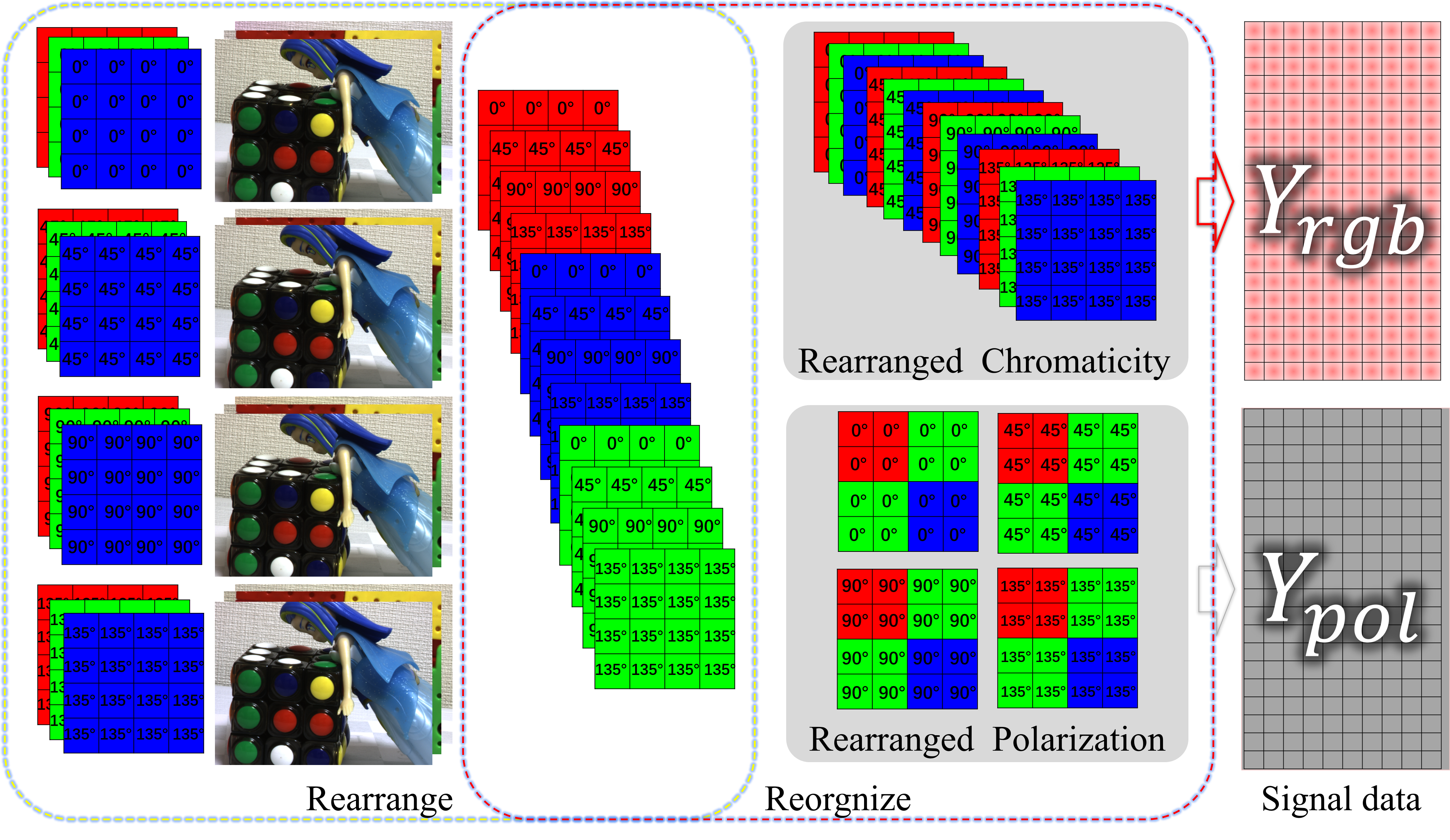}
		\end{center}
		\caption{The generation of signal data for chromatic dictionary and polarimetric dictionary.}
		\label{signal}
	\end{figure}
	\begin{figure*}[t]
		\begin{center}
			\includegraphics[width=1\linewidth]{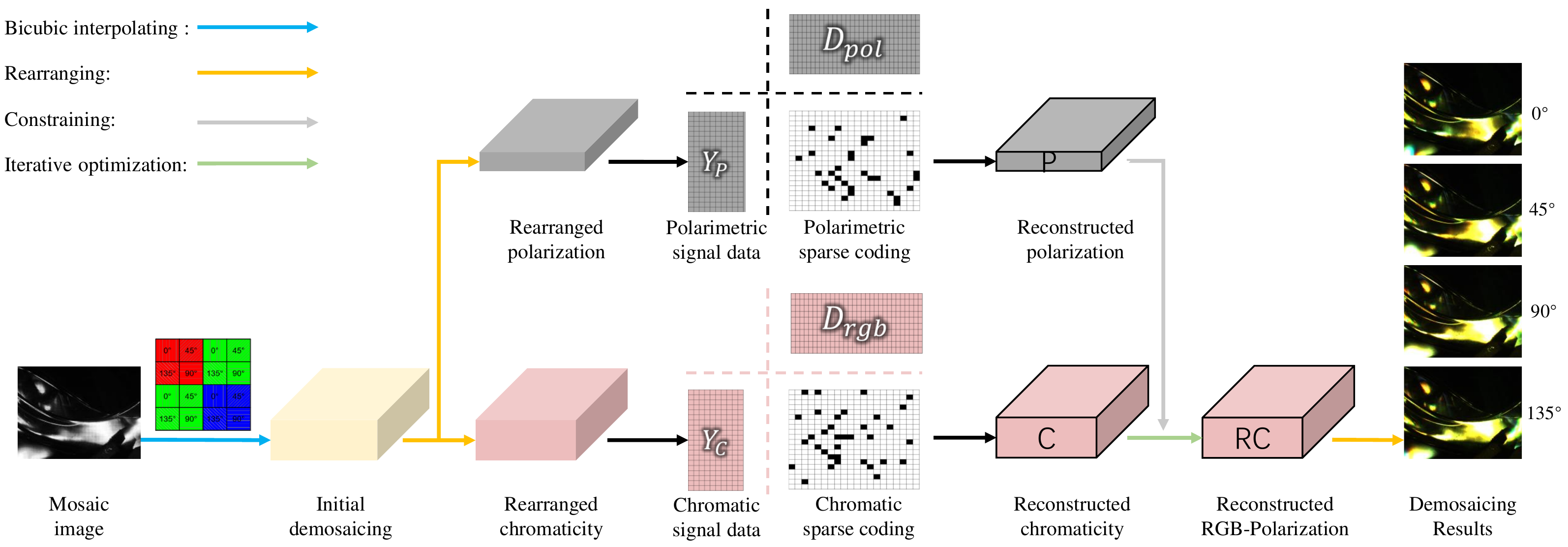}
		\end{center}
		\caption{The iterative optimization framework with learned dictionaries $\left\{\br{D}_{\theta}|\theta \in (pol, rgb) \right\}$.}
		\label{op}
	\end{figure*}
	
	\subsubsection{Generating the signal data}
	
	Before learning the dictionaries, we need to generate the corresponding signal data. The pipeline of signal data generation is illustrated in Fig.~\ref{signal}. We distribute the RGB-Polarization dataset into 12 channels ((r,g,b)*($0^\circ$, $45^\circ$, $90^\circ$, $135^\circ$)) based on the RGB-Polarization pattern. Then we concatenate four reorganized polarization channels into polarimetric signal data group with a size of m$\times$n$\times$4, as shown in \textit{Rearranged Polarization} of Fig.~\ref{signal}. To learn the polarimetric dictionary and sparse code, we divide the polarimetric signal data into small patches with a size of $4\times4$ and randomly choose 60000 of them. Then we lexicographically arrange the chosen patches as column vectors to generate the polarimetric signal data with a size of 64$\times$60000, denoted by $\br{Y}_{pol}$.
	
	We also need to generate the chromatic signal data for the chromatic dictionary. We can directly concatenate four full-resolution polarized color images from the RGB-Polarization dataset into a new form of data with a size of m$\times$n$\times$12. Then we divide this new-form data into patches and rearrange them to column vectors as the signal data. The size of generated signal data is 192$\times$60000, denoted by $\br{Y}_{rgb}$. 
	
	\subsubsection{Learning the dictionaries}
	
	After generating the signal data, we need to construct two over-complete dictionaries for chromatic and polarimetric dictionaries as initialization. The dictionary can be obtained by solving the following optimization problem:
	\begin{equation} \label{dictionary}
	{L(\br{D}_\theta, \br{X})} = 
	\arg \mathop{min} \limits_{\br{D}_\theta,\br{X}} \|\br{Y}_\theta - \br{D}_\theta \br{X}\|_F^2 + \lambda \|\br{X}\|_1.
	\end{equation}
	The number of atoms in $\br{D}_\theta$ is set to 256. The size of $\br{D}_{pol}$ and $\br{D}_{rgb}$ is 64$\times$256 and 194$\times$256. \br{X} is initial representation vector with size of 256$\times$60000. $\lambda$ is L1 norm regularization parameter and set to 0.0001 as shown in Tab.~\ref{comparison3}. According to K-SVD algorithm~\cite{aharon2006k,dumitrescu2017regularized}, the optimization is solved by updating one of the parameters with the other variables fixed. After the iteration, we will obtain the learned dictionaries $\left\{\br{D}_{\theta}|\theta \in (pol, rgb) \right\}$. The learned dictionaries will maintain the size and calculate the sparse coding of the signal data, generated from the input mosaic image.
	
	\subsection{Update P and C}\label{ini}
	After obtaining the polarimetric and chromatic dictionaries, we can calculate the corresponding sparse coding to estimate the desire results of \br{P} and \br{C}. By using the reconstructed polarization as a constrain, the main iterative optimization framework is illustrated in Fig.~\ref{op}. 
	
	\subsubsection{Initialization}
	
	Before the iteration, we need to initialize the input mosaic image. The size of the input mosaic image is m$\times$n$\times$1. As shown in the initialization of Fig.~\ref{op}, we reorganize the input mosaic image to 12 channels based on the RGB-Polarization pattern, which the size will be m$\times$n$\times$12. Then we use the Bicubic interpolation method to estimate the missing pixels, which is the preprocessing operation commonly used in the field of image super-resolution~\cite{dong2013nonlocally,dong2011image,mairal2009non,zhang2018sparse}. After the initial demosaicing, we need to generate the signal data $\br{Y} \in \left\{ P,C \right\}$ of the input. Similar to the strategy of generating signal data for dictionary construction, we rearrange the RGB information and polarimetric information to obtain the signal data $\br{Y} \in \left\{ P,C \right\}$, only with the size of input image. It should note that the initialization only applied in the first iteration. 
	\begin{figure*}[t]
		\begin{center}
			\includegraphics[width=1\linewidth]{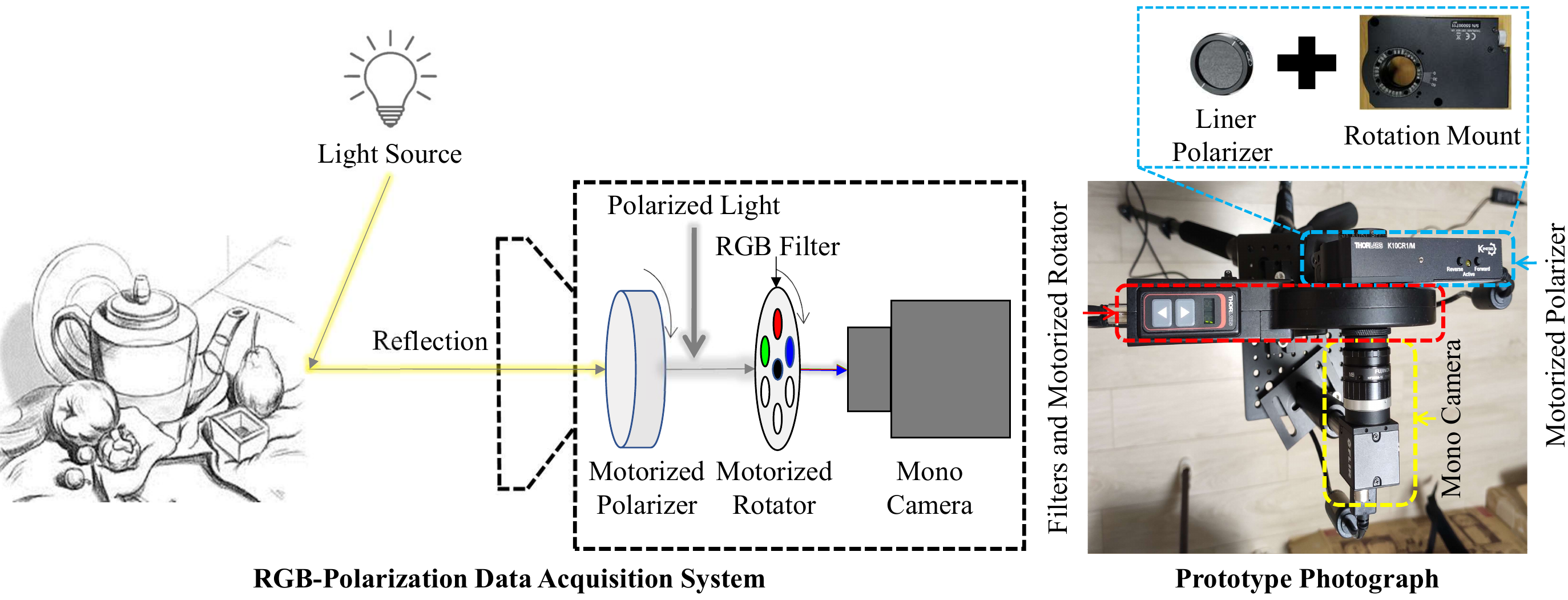}
		\end{center}
		\caption{The hardware implementation of our designed data acquisition system. The system consists of a motorized rotator with R,G,B filters, a mono camera, and a motorized polarizer with a rotation mount.}
		\label{acquisition}
	\end{figure*}
	
	\subsubsection{Calculating the sparse coding}As we mentioned before, the form of signal data $\br{Y} \in \left\{ P,C \right\}$ can be sparsely represented. The polarimetric and chromatic sparse coding $\left\{\br{X}_{\theta}|\theta \in (pol,rgb) \right\}$ are both sparse and low-rank. Furthermore, since the value of noise is sparse and non-negative, we also impose the non-negativity constraint on noise $\br{E}_\theta$. The calculation of polarization sparse coding $\br{X}_{\theta}$ can be expressed as an optimization function:
	\begin{equation} \label{p_code}
	\begin{aligned}
	&\arg \mathop{min} \limits_{\br{X}_\theta,\br{E}_\theta} rank(\br{X}_\theta) + \lambda\|\br{E}_\theta\|_0\\
	&s.t.\br{Y} = \br{D}_\theta\br{X}_\theta + \br{E}_\theta.
	\end{aligned} 	
	\end{equation}
	where $\lambda$ is the parameter used to balance the effect of the noise components. Minimizing the kernel norm is equivalent to minimizing the rank of the matrix. Therefore, to make the optimization traceable, we relax Eq.~\ref{p_code} by introducing $J$:
	\begin{equation} \label{p_code_1}
	\begin{aligned}
	&\arg \mathop{min} \limits_{\br{X}_\theta,\br{E}_\theta, J} \|J\|_* + \lambda\|\br{E}_\theta\|_0\\
	&s.t.\br{Y} = \br{D}_\theta\br{X}_\theta + \br{E}_\theta. \br{X}_\theta = J,
	\end{aligned} 	
	\end{equation}
	In this form, we can solve Eq.~\ref{p_code_1} by the well-known inexact ALM~\cite{lin2010augmented}. During the iteration, we use the OMP~\cite{tropp2007signal,pati1993orthogonal} to select the best matching atom from the corresponding dictionary to construct sparse approximation.
	
	Finally, the estimated $\br{P}$ and $\br{C}$ can be calculated by the dictionary (learned from the RGB-Polarization dataset) times the sparse coding (recovered from the output of initialization):
	\begin{equation} \label{PC}
	\begin{cases}
	{f(\br{P})} = {\br{D}_{pol}} \times \br{X}_{pol},\\
	{g(\br{C})} = {\br{D}_{rgb}} \times \br{X}_{rgb}.
	\end{cases}
	\end{equation}
	After obtaining the value $f(\br{P}), g(\br{C})$, we can update the $\br{P}$ and $\br{C}$ according to Eq.~\ref{sub2}. 
	
	\subsection{Update RP and RC}
	We fix $\br{P}$ and $\br{C}$ and then use the ADMM algorithm to minimize the formula Eq.~\ref{sub1} rather than a closed-form solution. It found that using optimization would yield better results. Then we can update $\br{RP}$ and $\br{RC}$.
	
	\subsection{Update $\br{S}_{pol}$ and $\br{S}_{rgb}$}
	Multipliers also need updating during each of the iteration. We firstly set the penalty parameters $\rho_{rgb}$ to 1.05 and $\rho_{pol}$ to 1.05. The update of scaled multipliers $\br{S}_{pol}$ and $\br{S}_{rgb}$ are as follows:
	\begin{equation} \label{S}
	\begin{cases}
	{\br{S}_{pol}^{k+1}} = \br{S}_{pol}^{k} + \br{RP}^{k+1} - \br{P}^{k+1},\\
	{\br{S}_{rgb}^{k+1}} = \br{S}_{rgb}^{k} + \br{RC}^{k+1} - \br{C}^{k+1}.
	\end{cases}
	\end{equation}
	
	\begin{algorithm}[t] 
		\caption{A Sparse Representation based Joint Demosaicing Method for Single-Chip Polarized Color Sensor.} 
		\label{Alg} 
		\begin{algorithmic}[1] 
			\Require Input the polarized color mosaic image $\br{I}$, $\rho_{rgb}$, $\rho_{pol}$, maxiter = 50, $\epsilon = 10^{-3}$
			\State Construct the Dictionaries $\left\{\br{D}_{\theta}|\theta \in (pol, rgb) \right\}$
			\State initialization~\ref{ini}; 
			\Repeat 
			\State Updating \br{P} and \br{C}; 
			\State Updating \br{RP} and \br{RC}; 
			\State Updating $\br{S}_{pol}$ and $\br{S}_{rgb}$; 
			\State Break: $\left\{ \left(\|\br{S}_{pol}^{k+1} - \br{S}_{pol}^{k}\| + \|\br{S}_{rgb}^{k+1} - \br{S}_{rgb}^{k}\|\right)  < \epsilon \right\}$
			\Until
			\Ensure $\br{RP}$, $\br{RC}^*$
		\end{algorithmic} 
	\end{algorithm}
	
	Overall, the proposed optimization of joint chromatic and polarimetric demosaicing is sumarized in Algorithm.~\ref{Alg}. After optimization, we will get the RGB-Polarization reconstruction $\br{RC}$ and arrange it into four full-resolution reconstructed polarized color images. As shown in Algorithm.~\ref{Alg}, we will obtain two results $\br{RP}$ and $\br{RC}$. However, the purpose of $\br{RP}$ is to constrain the reconstruction of polarimetric information during the optimization. It notes that the proposed method is based on the cross-optimization of chromaticity and polarization. More experimental results in the next section demonstrate the effectiveness and performance of the proposed method. 
	\begin{figure*}[htb]
		\begin{center}
			\includegraphics[width=1\linewidth]{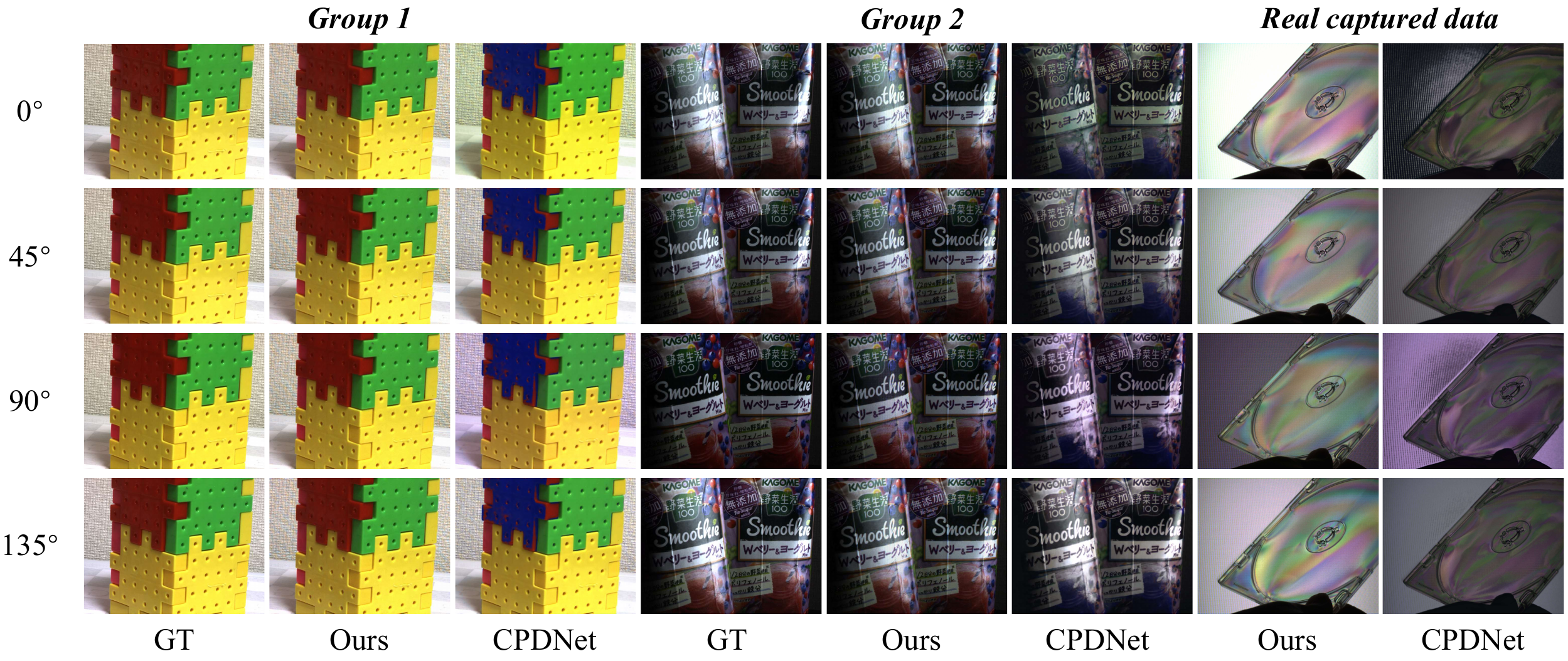}
		\end{center}
		\caption{Comparison results with the CPDNet trained by the Wen~\cite{wen2019convolutional}'s dataset.}
		\label{cpdnet}
	\end{figure*}
	
	\section{Experiments}\label{experiments}
	
	\subsection{Experiment Settings}\label{dataset}
	
	\subsubsection{Dataset}
	
	The proposed method requires the ground-truth polarized color images to learn the dictionary. Although Wen~\etal~\cite{wen2019convolutional} collected an RGB-Polarization dataset by adding a polarizer in front of a prism RGB camera. However, the prism RGB camera brings the chromatic aberration under different orientations of polarization. In addition, the CPDNet~\cite{wen2019convolutional} uses the captured RGB images of four polarization angles as ground truth and the corresponding synthesized mosaic images as input data to train the network. In this case, other sources of polarization, like illumination and birefringence, are not included in the CPDNet training data. To generate training pair in a way as CPDNet will greatly harm the performance on real scenes. We further prove our point in \secref{comE}.
	
	To address this issue, we design an RGB-Polarization data acquisition system and collect a new dataset for accurately constructing the dictionaries. The design schematic illustration and prototype of the data acquisition system are shown in Fig.~\ref{acquisition}. To separately capture the information from each degree of polarization, we rotate the polarizer by 0, 45, 90, and 135 degrees with a motorized rotator (Thorlabs K10CR1) to ensure that every set of images is aligned. Through each degree of polarization channels, we use an RGB filter with a motorized rotator (Thorlabs FW102C) to acquire the information of each R, G, B channel, respectively. Directly adding such a color filter in front of the light source is plausible, yet color mixture might occur when the scene contains fluorescence. Therefore, we add both the polarizer and the RGB filter in front of the camera. At last, we use a mono camera (BFS-U3-04S2m-cs) to capture each channel of polarized color images. In this way, we can avoid the unexpected chromatic aberration caused by the data acquisition system. The flickering effect of the fluorescent lamp will affect image acquisition. Though the extent of this impact is limited, we use a stabilized Xenon lamp to capture the dataset to alleviate blinking. In addition, we lengthen the exposure time to reduce this influence and keep the camera settings and position fixed throughout the imaging process.
	
	For our deep-learning free method, we collect the dataset includes 50 scenes which are composed of 600 one-channel 8-bit images with a size of 540$\times$720. The mosaic images are synthesized from the corresponding ground-truth images at four different polarization orientations and three RGB channels. The dataset consists of two groups, as shown in Fig.~\ref{cpdnet}. \br{\textit{Group 1}} includes 40 normal scenes that are captured under the unpolarized light source. We use 24 scenes of this group to construct the dictionaries. \br{\textit{Group 2}} includes 10 complex scenarios that are captured under the polarized light source. This group with polarized light aims to test whether our method can perform well under different sources of polarization, like illumination and birefringence. Our dataset contains multiple sources of polarization, which is completely different from the dataset of CPDNet.
	
	\subsubsection{Evaluation Metrics}
	
	The well-known evaluation metrics (PSNR, structural similarity index (SSIM)~\cite{wang2004image}, and color accuracy (CA)~\cite{wen2019convolutional}) are used to measure the accuracy of RGB information between the reconstructed image and the corresponding ground truth. We also need to concern the polarimetric information for the polarized color sensors. In this regard, we introduce the Stokes vector (S0), the degree of linear polarization (DoLP~\cite{gao2013gradient}) and angle of polarization (AoP~\cite{gao2013gradient}) to measure the accuracy of reconstructed polarimetric parameters. All these metrics are calculated by averaging four reconstructed images.
	
	\subsection{Comparative Experiments} \label{comE}
	\renewcommand{\arraystretch}{1.5} 
	\begin{table*}[htp]  
		\centering  
		\caption{The results of comparisons. CPDNet* is trained by Wen~\cite{wen2019convolutional}.}  
		\label{comparison1}  
		\begin{tabular}{|c|c|c|c|c|c|c|c|c|c|}  
			\hline  
			Method&Bilinear&Bicubic&RDN&VDSR&Zhang~\cite{zhang2018sparse}&CPDNet*&CPDNet&Qiu~\cite{qiu2019polarization}&ours\cr\hline  
			\hline  
			PSNR&25.860&26.602&22.093&29.386&29.341&20.336&27.855&29.660&$\boldsymbol{31.673}$\cr\hline  
			SSIM&0.868&0.880&0.780&0.932&0.927&0.665&0.873&0.932&$\boldsymbol{0.944}$\cr\hline  
			CA&21.492&21.197&19.688&22.133&21.063&17.4867&21.127&22.850&$\boldsymbol{23.318}$\cr\hline  
			S0&20.513&21.711&20.051 &24.386&26.021&15.368&25.733&25.264&$\boldsymbol{27.536}$\cr\hline  
			DoLP&24.265&24.321&15.099&25.243&25.491&24.374&25.240&$\boldsymbol{28.175}$&26.306\cr\hline  
			AoP&16.648&16.555 &13.573&16.501&17.324&13.608&14.968&17.447&$\boldsymbol{17.652}$\cr\hline  
			\hline  
		\end{tabular}  
	\end{table*}  
	
	Firstly, the proposed method is compared with the baseline interpolation algorithms such as Bicubic and Bilinear that have been proven effective in the image demosaicing processing field. Since the polarized color camera has just been released in a short time, as far as we know, only Wen~\etal\cite{wen2019convolutional} propose the CPDNet for this very task. To compare with the performance of the CPDNet, we retrain the network with the dataset, which is used to construct our dictionaries. To prove the accuracy of our dataset, we also directly use the trained CPDNet to perform on our test dataset. In order to verify the effectiveness of our approach, we also make comparisons with more state-of-art methods, include deep learning-based methods and sparse representation-based methods. 
	
	For deep learning-based method, we make comparative experiments with VDSR~\cite{kim2016accurate} and RDN~\cite{zhang2018residual}. Unlike super-resolution, our work focuses on a new filter array. It takes one channel polarized color mosaic image as input and recovers the other 11 channels ((r,g,b)*($0^\circ$, $45^\circ$, $90^\circ$, $135^\circ$)) for each pixel. To make fair comparisons with VDSR~\cite{kim2016accurate} and RDN~\cite{zhang2018residual}, we only modify the input and output as four polarized color images and keep their architecture unchanged. As shown in Tab.~\ref{comparison1}, the proposed method outperforms other methods in PSNR, SSIM, and Color Accuracy~\cite{wen2019convolutional}.
	
	The experimental results demonstrate that the existing methods cannot recover the RGB and polarimetric information as good as the proposed method. The deep learning-based methods will estimate the polarimetric information by all data which includes RGB information. In addition, the performance of deep learning-based methods also rely on the size of dataset. As a result, RDN performs even worse than the baseline interpolation methods on the evaluation metrics of polarization. Though the CPDNet uses Stokes parameters as a loss function to constrain the error of the polarization, the polarimetric and RGB information will still be mixed, and it is hard to exploit the feature separately during the training. 
	
	Since the size of our dataset is a limitation, overfitting may reduce the performance of the CPDNet. In this case, we also compare our method with the trained CPDNet by Wen~\cite{wen2019convolutional}. As shown in \textit{Group 1} of Fig.~\ref{cpdnet}, since the light in the general scene is unpolarized, the difference of color among four polarized color images should be slight. However, a clear alteration of color exists in the results of CPDNet. On the contrary, the data of \textit{Group 2} is captured under polarized illumination, which brings the different RGB images at four different polarization orientations. The comparison results show that our method still perform better than CPDNet. As a result, considering the deviation of synthesized data, CPDNet does not generalize well to real scenes. As shown in \textit{Real captured data} of Fig.~\ref{cpdnet}, CPDNet cannot accurately recover the polarimetric information and RGB information with birefringence polarization of the plastic disc cover. Different performances on real data distinguish our method from CPDNet most significantly. In addition, instead of fitting the synthesized dataset, our proposed method requires only a small amount of data to construct dictionaries which can provide satisfactory demosaicing results, without the cost of heavy training. 
	
	To compare with the sparse representation-based method proposed by Zhang~\etal\cite{zhang2018sparse}, due to the difference between PFA and RGB-Polarization pattern, we use each R, G, and B channel of the mosaic image as an input and concatenate the outputs in RGB form as a result. However, the PCA (principle component analysis~\cite{jolliffe1986principal}) with locally similar patches proposed by Zhang~\etal\cite{zhang2018sparse} requires a lot of calculation. Therefore, there is literally a limit on the size of the target image. In our case, the resolution of the input image is $540\times720$ which can not be processed by the maximum array size. For a fair comparison, we cut the test image into several small blocks with a size of $200\times200$ to reconstruct, and then recombine them together again. As shown in Tab.~\ref{comparison1}, the method proposed by Zhang~\cite{zhang2018sparse} cannot perform as well as the proposed method. In addition, Qiu~\etal~\cite{qiu2019polarization} newly propose a polarization demosaicing algorithm for color polarization focal plane arrays. However, their optimization model is based on the Stokes Parameters and aims to directly reconstruct the Stokes Parameters without paying attention to full-resolution polarized color images. As a result, shown in Tab.~\ref{comparison1}, although their method can obtain better results on the reconstructed DoLp images, it loses the details of the reconstructed full-resolution polarized color images.
	
\begin{figure*}[htbp]
		\begin{center}
			\includegraphics[width=0.96\linewidth]{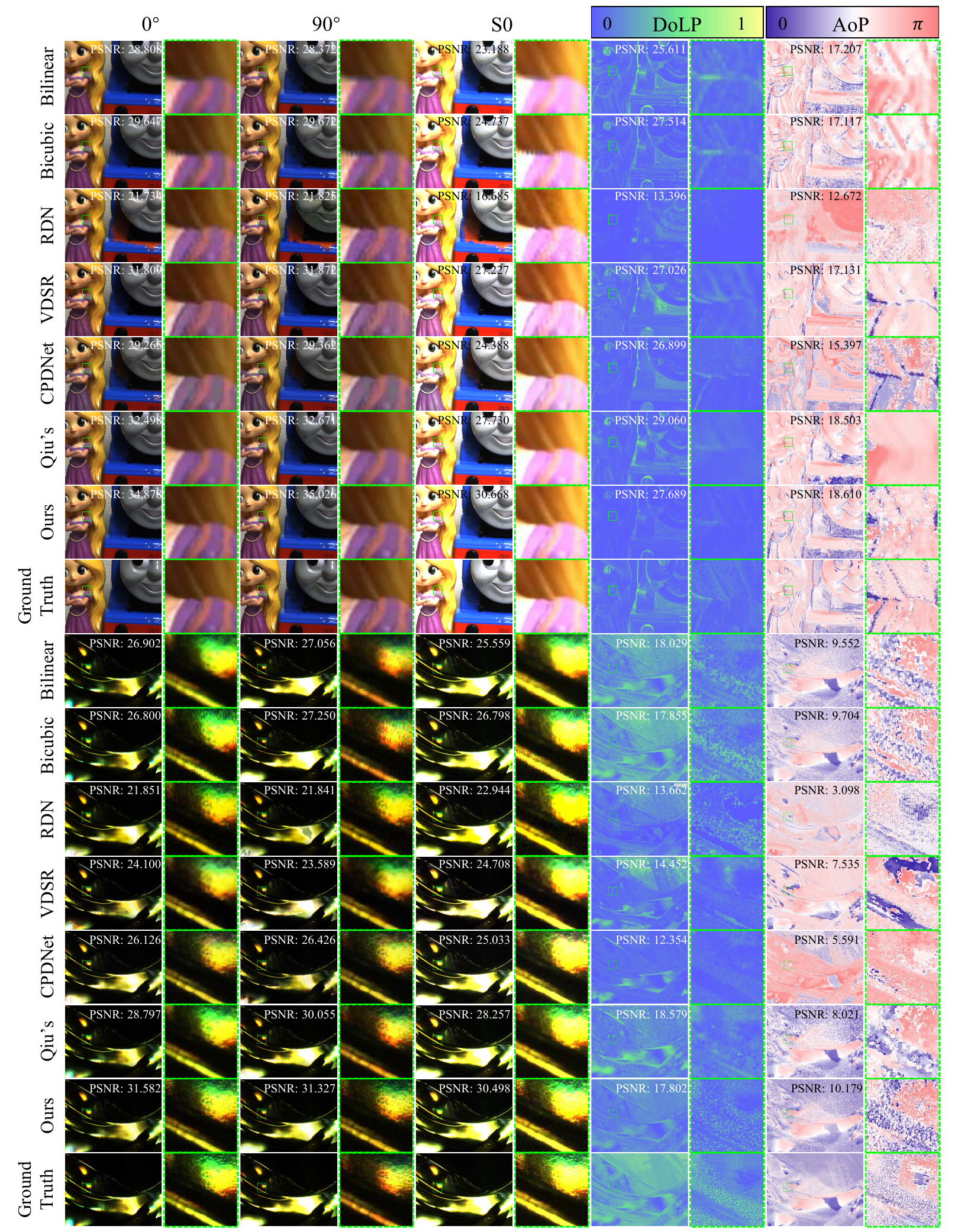}
		\end{center}
		\caption{We randomly choose two scenes in our dataset for experimental comparison. The upper part is a toy scene with unpolarized background illumination, and the lower part is a glass ornament scene with polarized background illumination. Zooming in will show more details.}
		\label{com_f}
	\end{figure*}

	We compare our method with others in two random scenes in our dataset, a toy scene with unpolarized background illumination and a glass ornament scene with polarized background illumination. The visualization of comparisons can be found in Fig.~\ref{com_f}, from which we can perceive that the proposed method outperforms all other algorithms. 
	\begin{figure*}[htb]
		\begin{center}
			\includegraphics[width=1\linewidth]{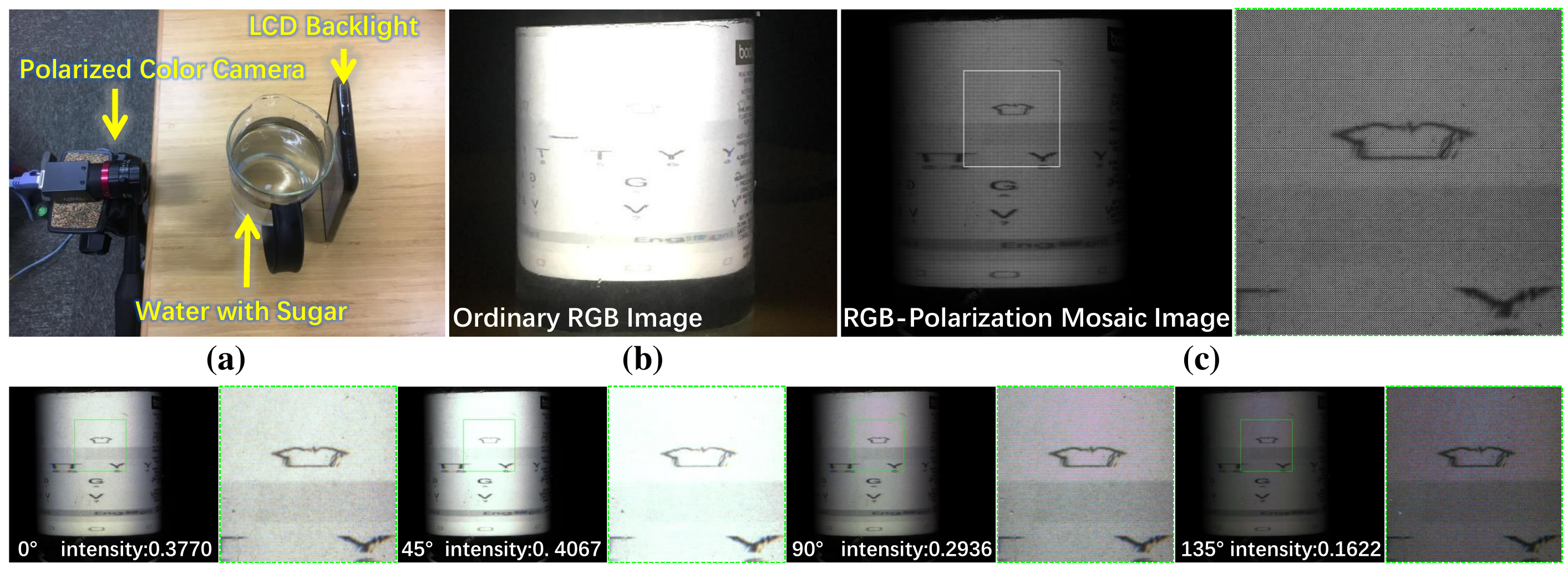}
		\end{center}
		\caption{Application: (a) Application experiment setup; (b) An image of a glass of water with sugar recorded by a normal RGB camera; (c) An mosaic image of the same scene recorded by a polarimetric RGB camera (Lucid PHX050S-QC); (d) Four full-resolution polarized color images reconstruct by our method.}
		\label{application}
	\end{figure*}
	
	\subsection{Controlled Experiments}
	\begin{table}[b]
		\renewcommand\tabcolsep{2.5pt} 
		\fontsize{9}{10}\selectfont
		\caption{The results of controlled experiments. }
		\begin{center}
			\begin{tabular}{l|cccccc}
				\toprule\hline
				Method & PSNR & SSIM & CA & S0 & DoLP & AoP \\
				\hline
				1-Dic           &27.742 &0.780 &21.032 &24.749 &22.354  &12.495 \\
				12-Dic          &26.812 &0.732 &20.413 &25.667 &21.512  &11.128 \\
				$\lambda=0.1$   &30.618 &0.934 &22.863 &26.741 &25.759  &16.567 \\
				$\lambda=0.01$  &31.444 &0.939 &23.167 &27.325 &26.323  &17.646 \\
				$\lambda=0.001$  &31.595 &0.943 &23.249 &27.420  &$\boldsymbol{26.448}$ &17.591\\
				$\lambda=0.0001$  &$\boldsymbol{31.673}$ &$\boldsymbol{0.944}$ &$\boldsymbol{23.318}$ &$\boldsymbol{27.536}$  &26.316 &$\boldsymbol{17.652}$\\
				$\lambda=0$  &31.641 &0.943	 &23.289 &27.472  &26.306 &17.629\\
				\hline\bottomrule
			\end{tabular}
		\end{center}
		\label{comparison3}
	\end{table}
	
	\subsubsection{Regularization Parameter} 
	
	As shown in Tab.~\ref{comparison3}, the L1 norm regularization parameter will affect the performance of our proposed method. When the value of the L1 norm regularization parameter is 0.0001, we can get better results in all evaluation metrics.
	
	\subsubsection{Different Dictionary}
	
	Constructing appropriate dictionaries plays an important role in sparse representation and low-level vision task. On the one hand, we can simply take the 12 channels of RGB-Polarization data to construct a dictionary~\cite{liu2010robust}. However, such a large dictionary is computationally expensive and consumes too much storage space. In addition, taking the entire data loses the detail of polarimetric and RGB information in the polarized color images. On the other hand, we can also construct 12 dictionaries based on each channel of the RGB-Polarization data. Since there is no interdependent correlation of each channel considered during the demosaicing process, the reconstruction results are even worse. The results in Tab.~\ref{comparison3} verify our the point of view. 
	
	\subsection{Qualitative Experiments}\label{realimage}
	
	\subsubsection{Application}
	\begin{figure}[!hb]
		\begin{center}
			\includegraphics[width=1\linewidth]{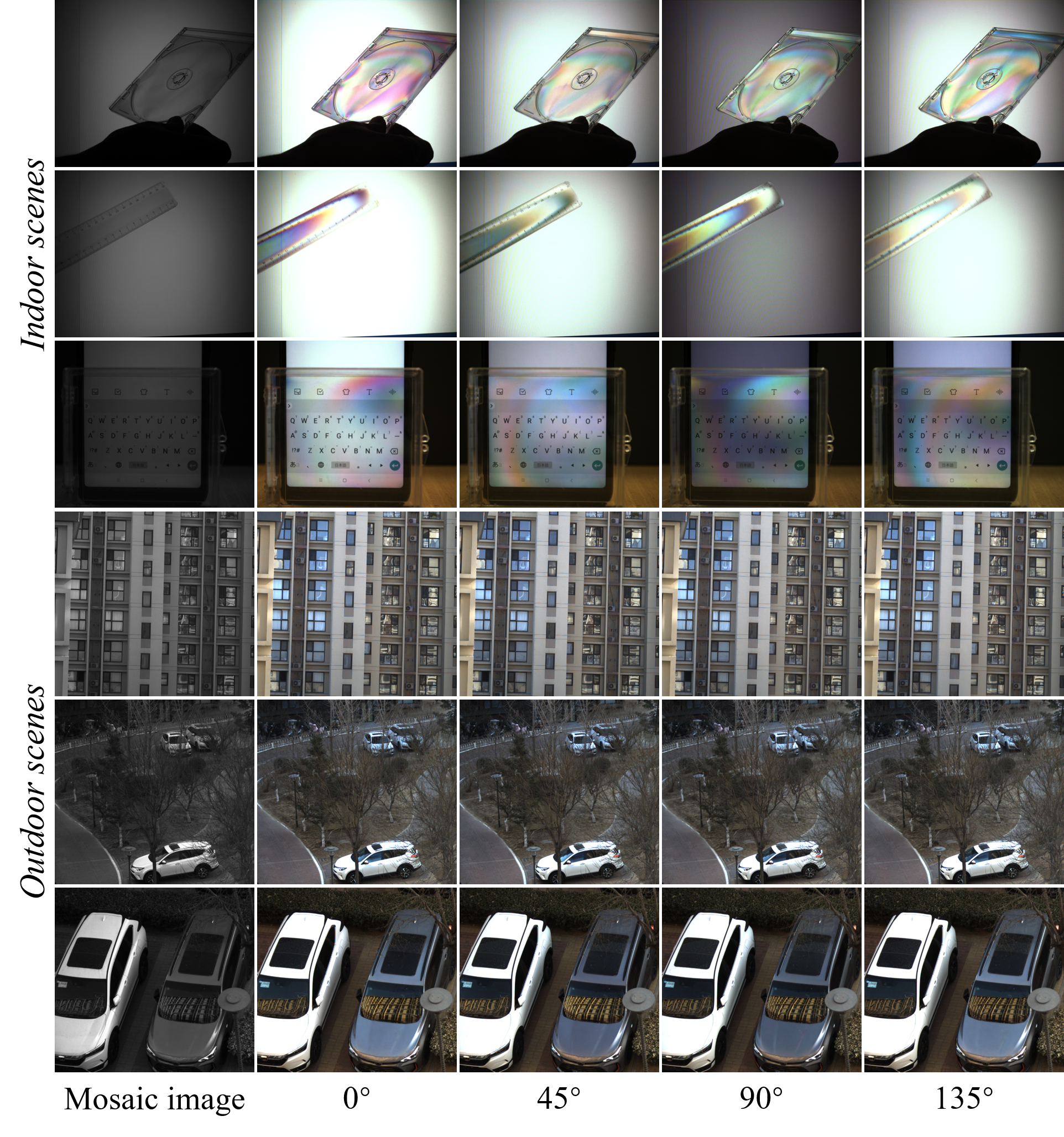}
		\end{center}
		\caption{Experiment results in the real scene.}
		\label{real}
	\end{figure}
	
	We conducted an application experiment to show the importance and effectiveness of joint chromatic and polarimetric demosaicing. In industry, the measurement of the concentration of sugar in a solution considers not only the effect of the solution under the polarized light but also many rigorous calibrations. This calibration operation is beyond our knowledge scope now. However, the observation of the shifted polarization angle is indeed one of the crucial steps in which our joint demosaicing method is critical. 
	
	The traditional operation of rotating the polarizer not only causes deviation but also makes the experiment process cumbersome. As shown in Fig.~\ref{application}, the ordinary RGB image taken with a normal camera can not display the color of polarized light that passes through a sugar solution. On the contrary, instead of shifting the polarization angle, the polarized color camera can directly capture the polarized color mosaic image by one snapshot. After processing along with the mosaic image by the proposed method, we can obtain \br{four polarization images with RGB information}. The ratio (calculated by RGB images) of known polarization angles ($0^\circ$, $45^\circ$, $90^\circ$, and $135^\circ$) can help to calculate the concentration of sugar. 
	
	\subsubsection{Real Scene}
	
	Polarimetric imagery can show the particular direction of the oscillation of the electric field described by the light. The ordinary RGB image taken with a normal camera can not display the color of polarized light. After the process of joint chromatic and polarimetric demosaicing, we can directly observe the color of the light in different angles of polarization orientation. To demonstrate the performance of our method in the real scene, we have captured images of a plastic box and a ruler in front of an LCD monitor. In addition, we also capture an image of the scene with the polarized light, which is the light of the screen that passes through a plastic box. To prove the performance of the outdoor scenes, we also conduct some experiments in more complex outdoor scenarios. As shown in Fig.~\ref{real}, the reconstruction results of the real captured scenes can achieve outstanding quality. 
	
\begin{figure}[htb]
		\begin{center}
			\includegraphics[width=1\linewidth]{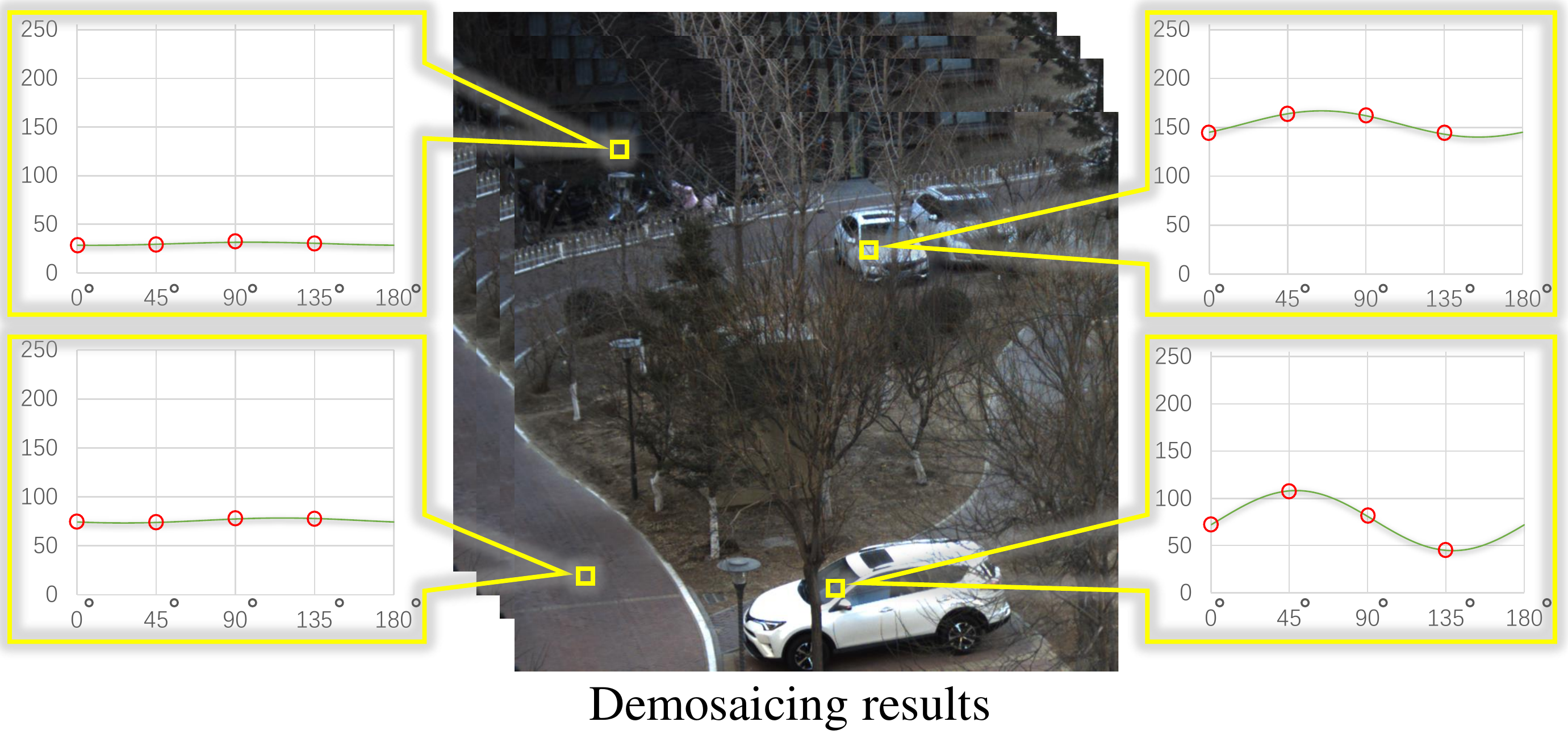}
		\end{center}
		\caption{The x-axis represents the degree of the polarizer ($0^\circ\sim180^\circ$). The y-axis represents the irradiance of the selected pixel. We select four areas (yellow boxes) of the scene and calculate their intensity fluctuation curve of polarization orientation in the red channel.}
		\label{plot}
	\end{figure}

	To further verify the effectiveness of the proposed algorithm, we randomly choose some pixels in the real captured outdoor scene and calculate their intensity fluctuation curve of polarization orientation. As shown in Fig.~\ref{plot}, the light from the cover and windows of the car has a certain amplitude under different polarization. Different from these strong polarized materials, the reflected light of the wall and ground surface is almost unpolarized. The plots show that the polarization information is accurately reconstructed, which demonstrates the effectiveness of the proposed algorithm.
	
	\section{Conclusion}\label{conclution}
	
	In this paper, we present a new joint chromatic and polarimetric demosaicing method, which can provide satisfactory demosaicing results without the cost of heavy training. To construct the dictionaries, we build a data acquisition system by using a mono camera, which is equipped with an RGB filter and the mechanically motorized linear polarizer. Based on the RGB-Polarization array pattern, this method transforms the joint demosaicing problem into an energy function that can be solved by the custom ADMM optimization scheme with sparse coding. The experimental results demonstrate our proposed solution is effective and practical. As we have shown in the experiments on the real image, joint chromatic and polarimetric imaging can be a benefit to computer vision tasks. In the future, we will continue yielding improvements in imaging quality and optimizing runtime. Other future work aspects include demosaicing optimization and specific applications such as 3D modeling, intrinsic image decomposition, specularity removal, object detection and recognition, and so on.
	
	\ifCLASSOPTIONcaptionsoff
	\newpage
	\fi
	
	
	
	%
	%
	%
	\bibliographystyle{IEEEtran}
	\bibliography{egbib}
	
	%

	\begin{IEEEbiography}[{\includegraphics[width=1in,height=1.25in,clip,keepaspectratio]{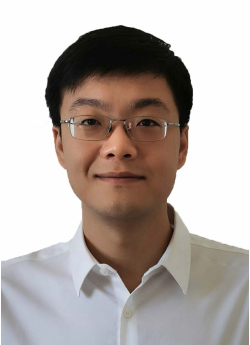}}]{Sijia Wen} is currently pursuing the Ph.D. degree in Technology of Computer Application with the State Key Laboratory of Virtual Reality Technology and Systems, School of Computer Science and Engineering, Beihang University. His research interests include computer vision and polarization-based vision analysis.
	\end{IEEEbiography}
	
	\begin{IEEEbiography}[{\includegraphics[width=1in,height=1.25in,clip,keepaspectratio]{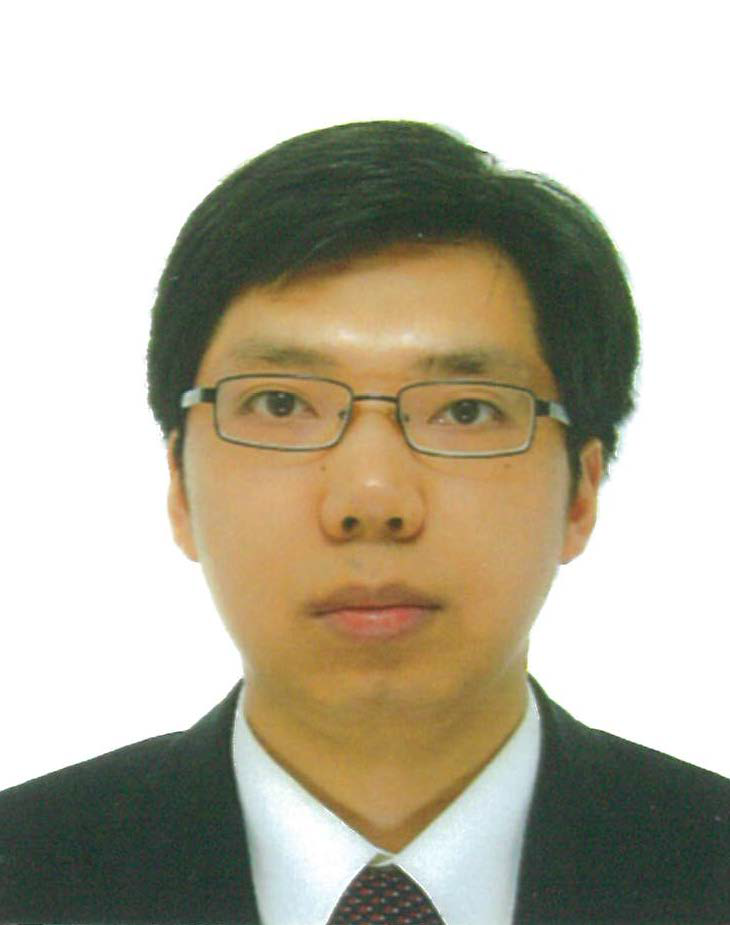}}]{Yinqiang Zheng} received his Bachelor degree from the Department of Automation, Tianjin University, Tianjin, China, in 2006, Master degree of engineering from Shanghai Jiao Tong University, Shanghai, China, in 2009, and Doctoral degree of engineering from the Department of Mechanical and Control Engineering, Tokyo Institute of Technology, Tokyo, Japan, in 2013. He is currently an associate professor in the University of Tokyo, Japan. His research interests include image processing, computer vision, and mathematical optimization.
	\end{IEEEbiography}
	
	\begin{IEEEbiography}[{\includegraphics[width=1in,height=1.25in,clip,keepaspectratio]{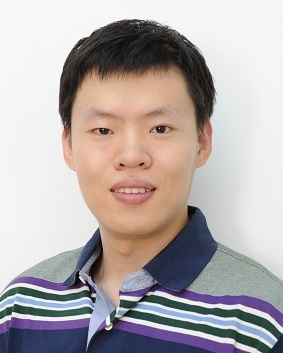}}]{Feng Lu} received the B.S. and M.S. degrees in automation from Tsinghua University, in 2007 and 2010, respectively, and the Ph.D. degree in information science and technology from The University of Tokyo, in 2013. He is currently a Professor with the State Key Laboratory of Virtual Reality Technology and Systems, School of Computer Science and Engineering, Beihang University. His research interests include computer vision, human-computer interaction and augmented intelligence.
	\end{IEEEbiography}
	
	%
	%
	
	
	

\end{document}